\documentclass[10pt,twocolumn,letterpaper]{article}

\usepackage{cvpr}

\usepackage[utf8]{inputenc} 
\usepackage[T1]{fontenc}    
\usepackage{url}            
\usepackage{booktabs}       
\usepackage{amsfonts}       
\usepackage{nicefrac}       
\usepackage{microtype}      
\usepackage[table]{xcolor}  
\usepackage{amsmath}        

\usepackage{graphicx}

\newcommand*\numcircled[1]{\raisebox{.5pt}{\textcircled{\raisebox{-.9pt} {#1}}}}

\definecolor{yellow25}{RGB}{254,237,204}
\definecolor{red25}{RGB}{240,195,206}

\usepackage[pagebackref,breaklinks,colorlinks]{hyperref}

\usepackage[capitalize]{cleveref}
\crefname{section}{Sec.}{Secs.}
\Crefname{section}{Section}{Sections}
\Crefname{table}{Table}{Tables}
\crefname{table}{Tab.}{Tabs.}

\def\cvprPaperID{6576} 
\def\confName{CVPR}
\def\confYear{2023}

\title{Just a Matter of Scale? Reevaluating Scale Equivariance\\in Convolutional Neural Networks\\[-0.5em]}
\author{
Thomas Altstidl$^1$ \, An Nguyen$^1$ \, Leo Schwinn$^1$ \, Franz K\"oferl$^1$ \, Christopher Mutschler$^2$\\ Bj\"orn Eskofier$^1$ \, Dario Zanca$^1$ \\
$^1$ FAU Erlangen-N\"urnberg, Germany \, $^2$ Fraunhofer IIS, Germany \\[-.25em]
{\tt\small \{thomas.r.altstidl,an.nguyen,leo.schwinn,franz.koeferl,bjoern.eskofier,dario.zanca\}@fau.de} \\[-.25em]
{\tt\small christopher.mutschler@iis.fraunhofer.de}\\[-0.5em]
}

\begin{document}

\maketitle

\begin{abstract}
  The widespread success of convolutional neural networks may largely be attributed to their intrinsic property of translation equivariance. However, convolutions are not equivariant to variations in scale and fail to generalize to objects of different sizes.
  Despite recent advances in this field, it remains unclear how well current methods generalize to unobserved scales on real-world data and to what extent scale equivariance plays a role.
  To address this, we propose the novel Scaled and Translated Image Recognition (STIR) benchmark based on four different domains. Additionally, we introduce a new family of models that applies many re-scaled kernels with shared weights in parallel and then selects the most appropriate one.
  Our experimental results on STIR show that both the existing and proposed approaches can improve generalization across scales compared to standard convolutions. We also demonstrate that our family of models is able to generalize well towards larger scales and improve scale equivariance. Moreover, due to their unique design we can validate that kernel selection is consistent with input scale.
  Even so, none of the evaluated models maintain their performance for large differences in scale, demonstrating that a general understanding of how scale equivariance can improve generalization and robustness is still lacking.
\end{abstract}

\section{Introduction}

Convolutional neural networks (CNNs) have become a widely used machine learning method across various domains. Up until recently, they have held the best performance in the ImageNet Large Scale Visual Recognition Challenge (ILSVRC), which is an image classification challenge with 1000 classes \cite{deng_imagenet_2009,russakovsky_imagenet_2015}. The translation equivariance of (discrete) convolutions is one of the key factors contributing to their success \cite{mouton_stride_2020}. By applying the same weights to each area of an image, the learned patterns will be independent of the location. This naturally fits to images, where a change in viewpoint seldomly leads to a change in interpretation.

\begin{figure}
    \centering
    \includegraphics[scale=0.9]{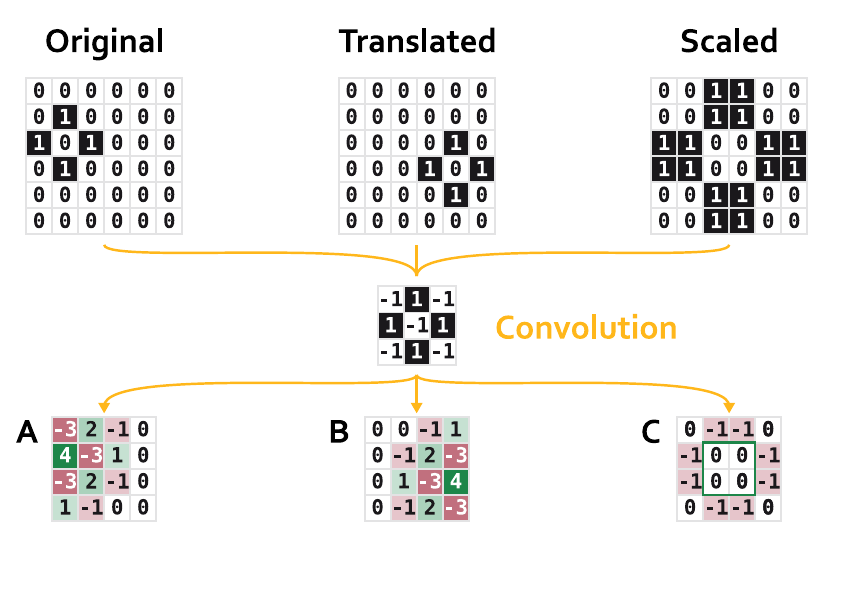}
    \vspace{-5mm}
    \caption{Result of applying a convolution on a translated and scaled variant of the original. In both the original (\textbf{A}) and translated (\textbf{B}) image the pattern is detected (see fours activated in green). This is due to the translation-equivariant property of convolutions. In the scaled (\textbf{C}) image the pattern is not detected (see framed zeros) due to the different pixel coverage.\\[-2em]}
    \label{fig:intuition}
\end{figure}

\begin{figure*}[t!]
    \centering
    \includegraphics[scale=0.9]{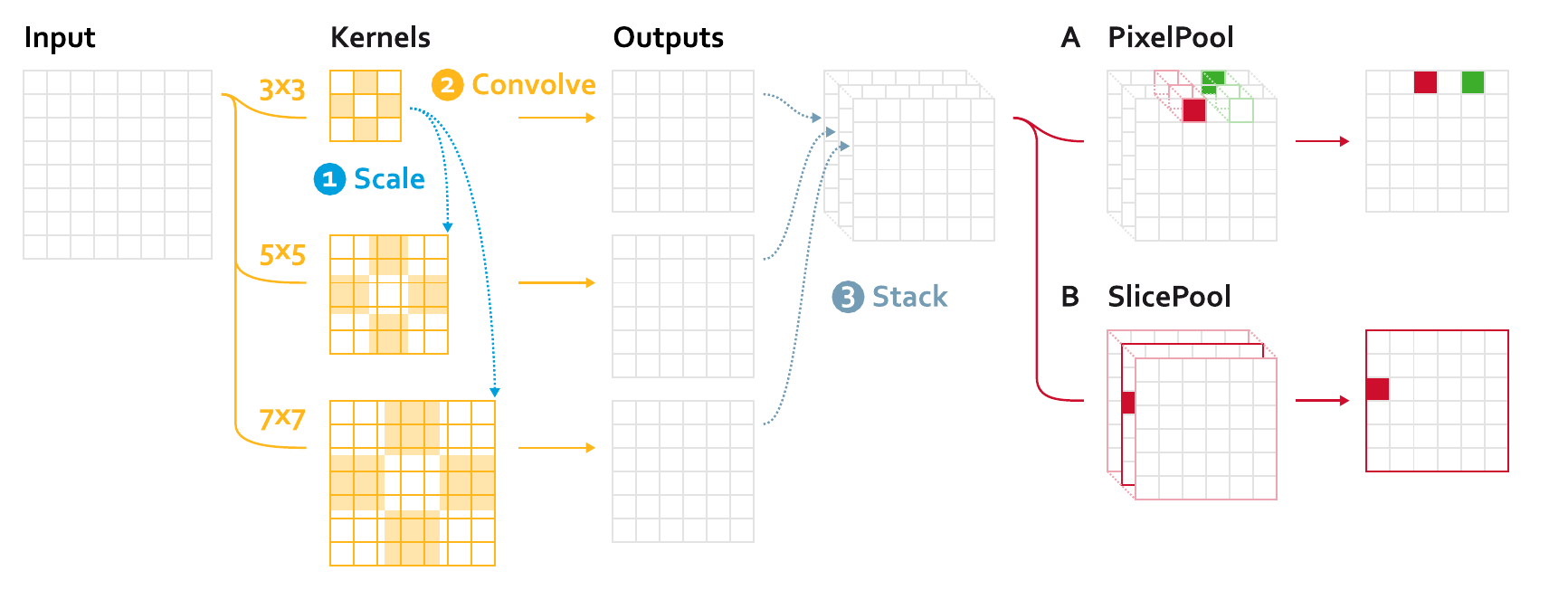}
    \caption{Overview of the scaled convolutional layer and subsequent pooling/convolution. \textbf{1} The base kernel (here $3 \times 3$ is up-scaled and interpolated in increments of two. \textbf{2} Each kernel is applied to the common input. \textbf{3} The outputs or feature maps of each kernel are stacked. After stacking two options are investigated. \textbf{A} Performing maximum pooling across the scale domain for each individual pixel. \textbf{B} Searching for the maximum across all domains and selecting the corresponding slice.}
    \label{fig:concept}
\end{figure*}

Nevertheless, the same cannot be said for other transformations, such as scaling, rotation, or reflections. In this work, we focus on the scale to reduce the overall scope, although our line of argumentation is equally applicable to others. \Cref{fig:intuition} illustrates our intuition. The given convolution kernel explicitly detects the plus-like pattern within the original image. Since convolutions are translation-equivariant, i.e., the result will be the same, save for changes due to translation, the same kernel also detects the pattern in the translated version. However, when scaling the pattern, here by a factor of $2$, the kernel does not provide any detection response. Convolutions are thus not scale-equivariant, and CNNs need to learn different kernels for different scales.

Ideally, pattern detection on images should perform equally well regardless of input scale. Numerous attempts have been made to address this. A common engineering solution is data augmentation \cite{liu_efficient_2018}. Some introduce architectural changes that explicitly model transformations \cite{jaderberg_spatial_2015,hinton_matrix_2018} or implicitly adapt to different scales \cite{van_noord_learning_2017,zhou_scale-transferrable_2018}. Others formulate local scale-equivariant kernels on the basis of theoretical \cite{ghosh_scale_2019,sosnovik_scale-equivariant_2020} or applicative \cite{kanazawa_locally_2014} considerations. Despite these recent advances, it remains unclear how suitable these are in practice due to a lack of comparative works.

Our contribution is twofold. First, we introduce the Scaled and Translated Image Recognition (STIR) benchmark. It is purposefully designed to cover a wide range of scales and contains four distinct domains, thus allowing us to probe scale generalization of specific models. Second, we introduce a new family of models that applies convolutions with repeatedly enlarged kernels based on a shared reference kernel, then pools by selecting among the resulting feature maps using different criteria. They are described in further detail in \cref{fig:concept} and \cref{sec:layer}. Their design allow us to explicitly analyse the selection of the pooling method with respect to the input scale.

\section{Related work}

Since convolutions are not inherently equivariant to scale, other mechanisms have been developed to ensure that CNNs maintain performance across varying scales.

One common method is data augmentation, where the training data is augmented with scaled versions of the original images \cite{liu_efficient_2018, boominathan_crowdnet_2016}. While this approach improves accuracy across scales, it is reliant on learning additional kernels and representative training data to capture variances in scale.

The \textit{Spatial Transformer} addresses this problem by using an auxiliary network that proposes and applies image transformations towards a common reference frame \cite{jaderberg_spatial_2015}. By doing so, recognition rates can be improved \cite{arcos-garcia_deep_2018}, but this transfers the problem of equivariance to the auxiliary CNNs. It has also recently been shown that they only have a limited capability to achieve invariance to scale\cite{finnveden_understanding_2021}.

The idea of the \textit{Ensemble} or \textit{Committee} originates from image segmentation \cite{farabet_learning_2013}, where the input image is resized multiple times and then processed using multiple equally structured CNNs. For pyramid generation, both Gaussian \cite{van_noord_learning_2017} and Laplacian \cite{farabet_learning_2013} have been used. Features are later concatenated \cite{farabet_learning_2013,shen_multi-scale_2015}, pooled \cite{van_noord_learning_2017,chen_attention_2016} or aggregated via attention mechanisms \cite{chen_attention_2016}. Some share weights across CNNs \cite{shen_multi-scale_2015}, while others don't \cite{van_noord_learning_2017}. There are also variants that scale kernels instead of images \cite{xu_scale-invariant_2014}. Foveated processing based on scale channels has also been evaluated theoretically and empirically for its ability to generalize to unseen scales \cite{jansson_exploring_2021,jansson_scale-invariant_2022}. No general consensus seems to have emerged about optimal configurations, but with shared weights, scale invariance of the CNNs across limited scales is possible.

Others rely on the structure of CNNs, where spatial pooling allows different depths to represent different scales. Feature maps are branched at various depths and later fused \cite{cai_unified_2016,nam_psi-cnn_2018,zhou_scale-transferrable_2018,sermanet_traffic_2011}. While this facilitates shortcuts for various scales, it does not try to directly enforce scale equivariance.

More recently, steerable filters have emerged as a mathematical framework that imposes limits on the kernel weights. Kernels are interpreted as a combination of basis functions, such as log-radial harmonics \cite{ghosh_scale_2019}, Hermite polynomials with Gaussian envelopes \cite{sosnovik_scale-equivariant_2020}, eigenfunctions of Dirichlet Laplacians \cite{zhu_scaling-translation-equivariant_2021} or discrete ones that minimize equivariance errors \cite{sosnovik_disco_2021}. However, this further restricts the patterns and can be difficult to implement in practice.

In addition, the combination of image resizing and subsequent maximum pooling has also been studied locally at the layer level \cite{kanazawa_locally_2014}. Extending that, a vector field formulation has been derived using the scale indices during maximum pooling \cite{marcos_scale_2018}. Another local method is based on scale-spaces operated on by dilated convolutions \cite{worrall_deep_2019}.


The most common dataset among all prior works is derived from MNIST \cite{ghosh_scale_2019,sosnovik_scale-equivariant_2020,kanazawa_locally_2014,marcos_scale_2018,jansson_exploring_2021} and measures either classification accuracy or error. Only few devise more detailed evaluations, such as on scale generalization \cite{jansson_exploring_2021,jansson_scale-invariant_2022} or equivariance errors \cite{sosnovik_scale-equivariant_2020,sosnovik_disco_2021,worrall_deep_2019}.

\section{Rethinking the convolution}

Let $f: \mathbb{Z}^2 \rightarrow \mathbb{R}^K$ be the input signal of a convolution as in \cite{cohen_group_2016}. The function $f$ maps a pixel coordinate $(x, y) \in \mathbb{Z}^2$ to a $K$-dimensional feature vector, with $K$ being the number of channels. In images, this input is usually subject to a wide range of transformations, including scaling, rotation, reflection, and skewing, among others. Limiting ourselves to a translation vector $(t_x, t_y) \in \mathbb{R}^2$ and a scaling factor $s \in \mathbb{R}$, such a transformation can be described by
\begin{equation}
    g(x, y) \simeq
    \begin{pmatrix}
    s & 0 & t_x \\
    0 & s & t_y \\
    0 & 0 & 1
    \end{pmatrix}
    \begin{pmatrix}
    x \\ y \\ 1
    \end{pmatrix},
\end{equation}
where $g$ maps (homogeneous) pixel coordinates and can be applied to any signal $f$ using $L_g f = f \circ g^{-1}$.

It is known that convolutions $\star$ with filters $\psi: \mathbb{Z}^2 \rightarrow \mathbb{R}^K$ are inherently equivariant to translation. That is, if $s = 1$ and thus no scaling is involved, the equation
\begin{equation}
    (L_g f) \star \psi = L_g (f \star \psi)
\end{equation}
holds. However, the same is not true for scales $s \neq 1$. To be scale-equivariant, a new type of operation $\star'$ is needed.

A standard convolution is equivariant to translation as it essentially performs a brute-force pattern search along the dimensions covered by $t_x$ and $t_y$. Intuitively, this should also extend to the scale $s$. The subsequent section will introduce our approach for performing a pattern search covering that dimension. Afterwards, different models are presented that incorporate this novel layer.

\subsection{Scaled convolutional layer}\label{sec:layer}

Our layer, which we'll subsequently call \textit{SConv2d}, builds upon the intuition that a convolution is essentially a brute force search across the two translation domains. It would thus be logical that the search should be extended to the scale domain. The chosen approach is summarized in \cref{fig:concept}.

A standard convolution is normally performed using a $k \times k$ kernel on a $n \times m$ input image or feature map. For simplicity, we assume that $n = m$. The presented layer operates similarly but builds multiple larger kernels based on the $k \times k$ base kernel.

Kernel sizes are progressively increased by steps of two to prevent alignment issues. Decreases in kernel size are avoided to prevent loss of information due to undersampling. Kernel generation is stopped once it is incompatible with the input, such that
\begin{equation}\label{eq:scales}
    s = \left\lfloor \frac{n - k}{2} \right\rfloor + 1
\end{equation}
kernels are generated in total. In the given example, the $3 \times 3$ base kernel is thus resized to $5 \times 5$ and $7 \times 7$ (\numcircled{1} in \cref{fig:concept}).

Most importantly, the number of parameters does not change compared to a normal convolution. Only the weights of the base kernel are learned, and the other kernels are derived via interpolation. The interpolation is bicubic, which has shown favorable performance.

Each generated kernel is applied individually to the common input using a standard convolution (\numcircled{2} in \cref{fig:concept}). The result is a set of output feature maps that, without loss of generality, shall be sorted by kernel size. These are then stacked such that an additional dimension is created (\numcircled{3} in \cref{fig:concept}). If necessary, the feature maps are zero-padded to conform to the largest size.

By producing an additional dimension representing the scale transformation, we handle scale equivariance similarly to translation in existing convolutions. Instead of only computing feature correlations for each $x$- and $y$-coordinate, we also compute those correlations for different scales. However, due to the added dimension, this structure cannot be integrated into existing architectures by itself.

Compared to existing models, it is closest to \cite{kanazawa_locally_2014} in that it operates locally, but instead of scaling inputs, it scales kernels as in \cite{xu_scale-invariant_2014}. It also makes use of more scales than in other related work and is followed by novel pooling methods.

\subsection{Proposed scale-equivariant models}\label{sec:proposed}

\begin{table}
    \centering
    \begin{tabular}{ccc}
        \toprule
        Standard & PixelPool & SlicePool \\
        \midrule
        \rowcolor{yellow25} $7 \times 7$ Conv2d & $7 \times 7$ SConv2d & $7 \times 7$ SConv2d \\
        \rowcolor{red25} -- & pixel & slice \\
        \multicolumn{3}{c}{ReLU} \\
        \rowcolor{yellow25} $7 \times 7$ Conv2d & $7 \times 7$ SConv2d & $7 \times 7$ SConv2d \\
        \rowcolor{red25} -- & pixel & slice \\
        \multicolumn{3}{c}{ReLU} \\
        \multicolumn{3}{c}{Global Pooling} \\
        \bottomrule
    \end{tabular}
    \caption{Overview of architectures with different pooling strategies. The first convolution uses 16 kernels, the second one 32. Yellow shading indicates convolution, red shading pooling.}
    \label{tab:models}
\end{table}

\begin{figure*}[t!]
    \centering
    \includegraphics[scale=0.9]{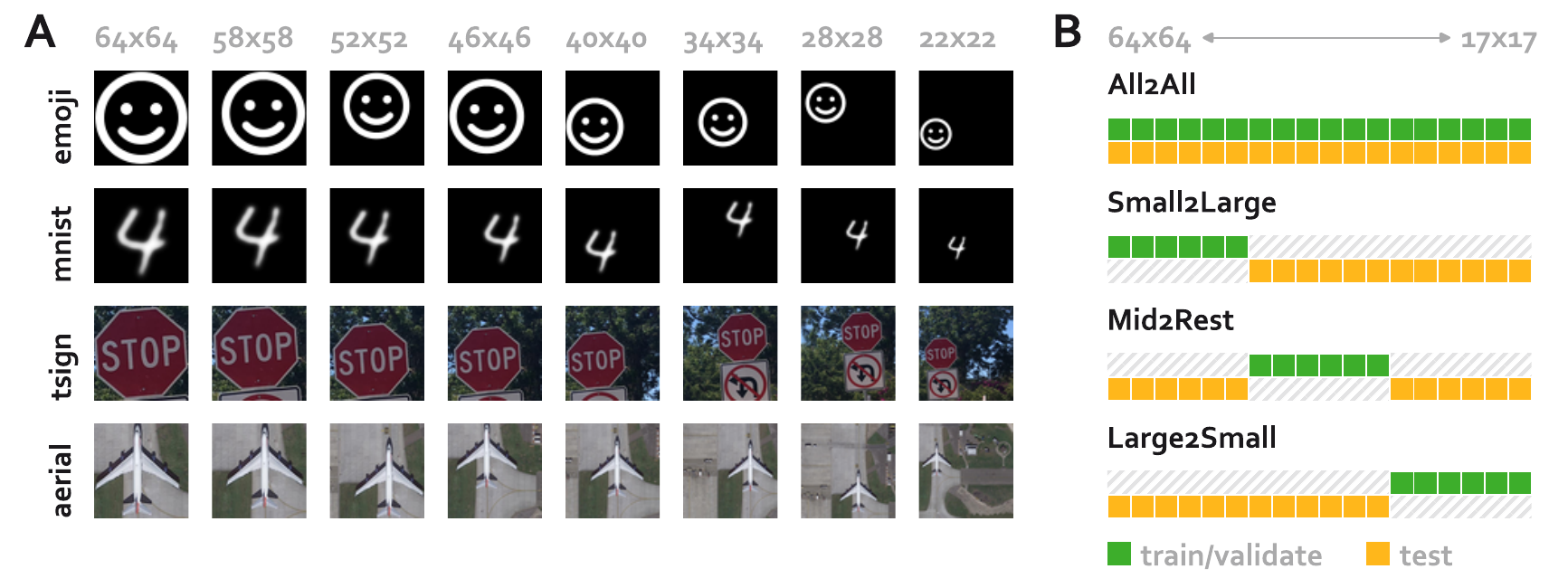}
    \caption{Overview of Scaled and Translated Image Recognition (STIR) dataset. \textbf{A} shows one example instance each of the four subsets emoji, mnist, tsign, and aerial. The first two are black and white while the other two are color images. \textbf{B} illustrates the different evaluation scenarios which differ in what scales are used for training and testing. Best viewed zoomed in and in color.}
    \label{fig:dataset}
\end{figure*}

We propose two different pooling strategies to adapt to the additional scale dimension. Both of them are based on variants of maximum pooling over the scale domain. The two pooling-based approaches are further visualized in \cref{fig:concept}, on the right-hand side. The resulting architectures and layers are summarized in \cref{tab:models}. All convolution layers make use of a $7 \times 7$ kernel to maintain comparable parameter counts.

\textbf{PixelPool} uses standard maximum pooling solely over the scale domain. In other words, its kernel size is $s \times 1 \times 1$, where $s$ is the number of scales produced by the scaled convolutions given in \cref{eq:scales}. Here, each spatial pixel is pooled independently of the others. This is also the approach followed by Kanazawa et al.~\cite{kanazawa_locally_2014}.

\textbf{SlicePool} aims for a more coherent response among pixels in the output. The maximum (of a single channel) is searched for across all scales and pixels. Let $(i_k, i_x, i_y)$ be the location of that maximum. Then the output will consist of all values at positions $(j_k, j_x, j_y)$ where $i_k = j_k$. By doing so, all spatial pixels of the output will be taken from the same scale dimension $i_k$.

\subsection{Reference models}\label{sec:reference}

To provide a fair evaluation, we implement seven architectures presented in related literature. They comprise a representative selection of the related methods introduced in earlier work. Where possible, the number of parameters was kept similar to our configurations.

\textbf{Standard} serves as the baseline CNN with no adaptations made to cover differing scales. Its configuration is listed in further detail in \cref{tab:models} and symmetric to proposed models.

\textbf{SpatialTrans} uses a single transformation network with two convolutional layers for warping the image \cite{jaderberg_spatial_2015} before performing classification based on the \textit{Standard} architecture. This incurs additional parameters compared to others.

\textbf{Ensemble} transforms the input image into a Gaussian pyramid with sizes $64 \times 64$, $32 \times 32$, and $16 \times 16$ \cite{van_noord_learning_2017}. These three different scales are processed by copies of the \textit{Standard} architecture with shared weights across columns. Features of different columns are aggregated by averaging \cite{van_noord_learning_2017,jansson_exploring_2021}.

\textbf{Xu} is directly based on the work of Xu et al. \cite{xu_scale-invariant_2014}. Multiple columns with different kernel sizes $k \times k$, with $k \in \{3, 5, 7, 9, 11\}$ process the input image in parallel. The kernel $7 \times 7$ serves as the reference and others are interpolated bilinear (downscaling) or nearest (upscaling). Features of different columns are aggregated using maximum pooling.

\textbf{Kanazawa} is directly based on the work of Kanazawa et al. \cite{kanazawa_locally_2014}. It operates locally at the layer level instead of the architecture level. Input images or feature maps are scaled by factors $1.26^e$, with $e \in \{-2, \ldots, 4\}$. The shared kernel is then applied on each, and features are subsequently maximum pooled in the scale domain.

\textbf{Hermite} is a scale-equivariant steerable (SES) model \cite{sosnovik_scale-equivariant_2020} that uses Hermite polynomials with Gaussian envelopes as base functions. We choose scale factors $s \in \{2.0, 2.52, 3.17, 4.0\}$ with a base kernel size of $7 \times 7$ to compute the applied convolutional kernels.

\textbf{Disco} also is part of the SES model family. It empirically optimizes a set of base functions to minimize discrete equivariance errors \cite{sosnovik_disco_2021}. We use the precomputed functions provided by the authors for scale factors $s \in \{1.0, 1.26, 1,59, 2.0\}$ with a base kernel size of $7 \times 7$.

\section{Datasets}\label{sec:data}

\begin{figure*}
    \centering
    \includegraphics[scale=0.9]{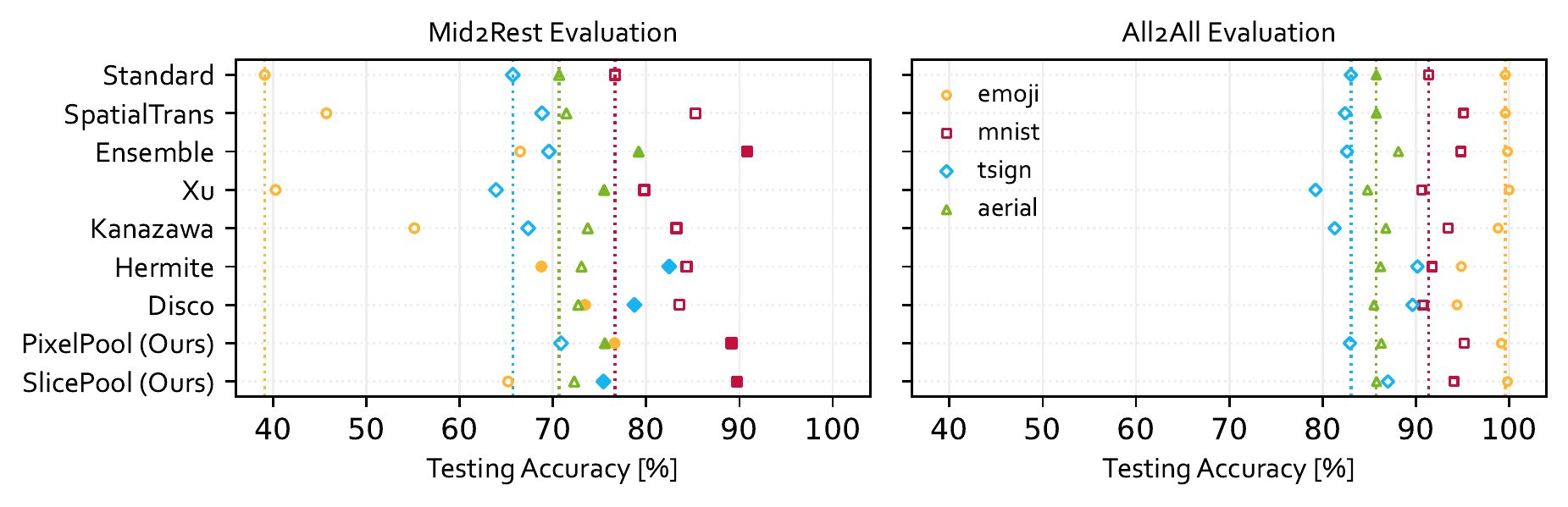}
    \caption{Mean testing accuracy on STIR across different datasets and models for two selected evaluation scenarios, both illustrated by \cref{fig:dataset}. Each accuracy is the average of 50 (emoji) or 25 (mnist, tsign, aerial) different seeds. Vertical dotted lines represent the baseline accuracy of the respective standard CNN. The best three models for each dataset are drawn filled, the remaining are unfilled.}
    \label{fig:performance}
\end{figure*}

While some benchmarks for evaluating scale equivariance exist, such as MNIST-scale \cite{kanazawa_locally_2014,ghosh_scale_2019}, they have their shortcomings as they are derived from transformations applied to existing images. They are thus subject to artifacts caused by aliasing or interpolation. In addition, they rarely cover a realistic spectrum of use cases which range from autonomous driving to remote sensing.

In light of these shortcomings, we devise a novel benchmark called Scaled and Translated Image Recognition (STIR) to experimentally validate scale generalization of a given model. It contains four different tasks, which classify emojis, handwritten digits, traffic signs and aerial scenes, respectively. The emojis are generated directly from vector graphics and are thus free of interpolation artifacts. The traffic sign and aerial scene images showcase typical applications that face a large variety of object scales and resolutions. All data is made freely available for download on Zenodo\footnote{\url{https://zenodo.org/record/6578038}}.

For each task, models need to classify a $64 \times 64$ pixels image into one of $n_c$ classes. A total of $n_i$ instances of each class are given in the training, validation and testing splits. Each instance is rendered at all incremental sizes between $17 \times 17$ and $64 \times 64$ for a total of 48 different scales. With the exception of the emojis, the instances are not overlapping.

\textbf{Emojis.} Due to their geometric simplicity, we settled on using the $n_c = 36$ emojis provided in the Font Awesome icon set \cite{fontawesome_download_2022}, which are available as Scalable Vector Graphics (SVG). They have been rendered directly from their vector representations, which describe the geometry with unlimited precision, and thus do not contain typical interpolation effects such as aliasing -- which are present for the others presented here.

\textbf{Handwritten Digits.} As it is a common reference in literature, we also provide resized handwritten digits derived from the original MNIST dataset \cite{lecun_gradient-based_1998}. We closely follow the approach of Janssen and Lindeberg \cite{jansson_scale-invariant_2022} by rescaling the original images using bicubic interpolation with scaling factors $s \in \{\frac{i}{28} \mid i = 17, \ldots, 64\}$. For each resized image we also apply Gaussian smoothing with $\sigma = \frac{7}{8} s$ and sharpen each pixel intensity $I$ using $I_{out} = \frac{2}{\pi} \arctan (0.02 I_{in} - 128)$. For each of the $n_c = 10$ classes, a set of $n_i = 50$ instances are randomly drawn from the training and testing splits, respectively.

\textbf{Traffic Signs.} The Mapillary Traffic Sign Dataset (MTSD) \cite{ertler_mapillary_2020} contains around 52k fully annotated street-level images. We only select labels where at least $n_i = 25$ instances for training, validation, and testing each are available at a size of $64 \times 64$ or above. This ensures all images are downsampled and results in a total of $n_c = 16$ traffic sign classes. Based on the annotated bounding boxes, each image is resized such that the traffic sign fits into a square $i \times i$ box, with $i = 17, \ldots, 64$. This ensures that models only need to scale in each direction equally.

\textbf{Aerial Objects.} The Dataset for Object Detection in Aerial Images (DOTA) \cite{xia_dota_2018} has around 403k annotated objects across 2,806 aerial images. We limit ourselves to color images derived from Google Earth. In addition, we constrain rotations to be below $3\deg$ as they are not a focus of the present work. Otherwise, we follow the same approach used for the traffic signs, selecting $n_c = 9$ classes where at least $n_i = 25$ instances of size $64 \times 64$ or larger exist. These are then each resized to sizes $i = 17, \ldots, 64$.

\section{Experiments}

\begin{figure*}
    \centering
    \includegraphics[scale=0.9]{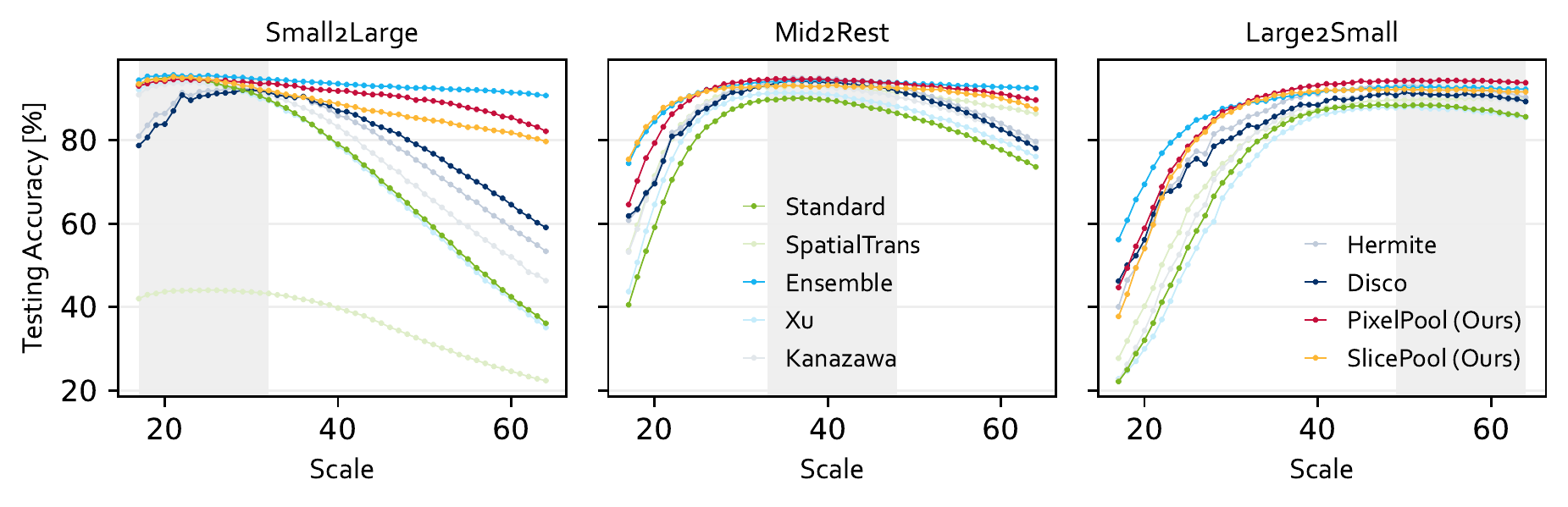}
    \caption{Mean testing accuracy per scale across different evaluation scenarios for the STIR handwritten digits dataset. Shaded areas represent scales used during training. Ideally, models would maintain their performance in the non-shaded areas and thus generalize to previously unseen scales. Best viewed zoomed in and in color.}
    \label{fig:generalization}
\end{figure*}

We perform extensive experiments to validate our assumptions about and the characteristics of different models with regards to scale equivariance. More specifically, we devise a set of three experimental techniques:

\begin{itemize}
    \item To show the generalization over scales, we use four evaluation scenarios described in \cref{sec:performance}. We train and evaluate on disjunct sets of scales, measuring performance in terms of classification accuracy.
    
    \item To empirically validate scale equivariance, we compare feature maps of specific layers using the $L^2$-norm for two differently scaled inputs.
    
    \item To show the effectiveness of the pooled representations, we analyze the correlation between the activated indices in the scale domain and the actual object scale.
\end{itemize}

Training of all models is done on a cluster where each node has two AMD Rome 7662 CPUs with 512 GB RAM along with four Nvidia A100 graphics card with 40 GB of memory. The implementation is done in PyTorch, with fixed global seeds. The optimizer is Adam \cite{kingma_adam_2017} with a learning rate of either $10^{-2}$ or $10^{-3}$ throughout all experiments. We initially train all models with both learning rates for five different seeds, then choose the best one based on the validation accuracy. Afterwards, a total of 50 seeds are trained for the emoji dataset and 25 for all other datasets.

\subsection{Model Performance}\label{sec:performance}

One of the core interests of this work is to investigate the ability of models to generalize to previously unseen scales. We thus devise four different evaluation scenarios that are visualized in \cref{fig:dataset}. In the \textit{All2All} scenario, all scales are used for both training and evaluation, representing an ideal case with no distribution shift in scales. The remaining three scenarios, \textit{Small2Large}, \textit{Mid2Rest} and \textit{Large2Small}, each train on just a third of all scales and test on the remainder. This allows us to evaluate performance both with and without differences in scale between training and testing data.

A summary of mean accuracies on the respective testing data is given in \cref{fig:performance}. A representative subset of evaluation scenarios, namely \textit{Mid2Rest} and \textit{All2All}, was selected. The interested reader is referred to supplementary \cref{fig:app_performance} where more detailed plots of all individual performances and all evaluation scenarios are given.

We find that there is not one model that outperforms all others. Performances on the \textit{All2All} scenario are generally high, as is to be expected since changes are mostly limited to different object instances rather than scales. Variability between datasets can likely be attributed mainly to inherently different complexities. The two SES models \cite{sosnovik_scale-equivariant_2020,sosnovik_disco_2021} are the best models on traffic signs and the worst models on emojis, thus showing highly dataset-dependent performance.

The \textit{Mid2Rest} scenario is more informative as the performances are indicative of generalization to smaller and larger scales. The only model that is among the top three models in three of the datasets is our \textit{PixelPool} model. The \textit{Ensemble}, the two SES models \cite{sosnovik_scale-equivariant_2020,sosnovik_disco_2021} and our \textit{SlicePool} model are within the best three for two datasets each. Lastly, \textit{Xu} \cite{xu_scale-invariant_2014} is among the top three for the aerial objects dataset. This suggests that our approach, SES models and ensemble-based models are equally effective at promoting generalization.

\subsection{Scale Generalization}\label{sec:generalization}

Testing accuracy itself is only a rough proxy for generalization. In particular, it remains unclear how performances change as the difference in scales increases. We thus also have a detailed look at how much accuracy degrades as scales are farther away from those used during training, as done previously by others \cite{jansson_exploring_2021,jansson_scale-invariant_2022}. To do so, accuracy is computed for each scale $s \in \{17, \ldots, 64\}$ and subsequently plotted.

\Cref{fig:generalization} shows these performance profiles for the handwritten digits subset of our STIR dataset. Others are given in supplementary \cref{fig:app_generalization}. Shaded areas represent scales used during training in the respective evaluation scenario. Models that generalize well are expected to exhibit horizontal lines, maintaining performance outside these shaded regions.

Most of the evaluated models' performances taper off towards the edges as scales deviate more and more from the training data. This is generally more pronounced on smaller scales than on larger ones. The three models that are the most stable are the \textit{Ensemble}, \textit{PixelPool}, and \textit{SlicePool}. This is followed by the two SES models, \textit{Hermite} and \textit{Disco}, both of which behave similarly. The remaining models perform comparably to the CNN baseline \textit{Standard}.

\subsection{Feature Map Similarity}\label{sec:equivariance}

\begin{figure}
    \centering
    \includegraphics[scale=0.9]{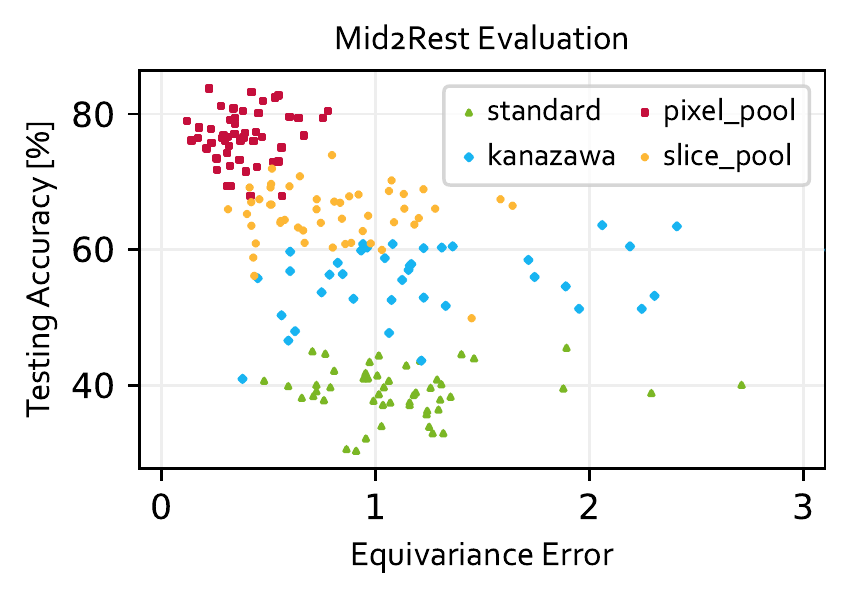}
    \caption{Measurement of equivariance error versus test accuracy on the emoji dataset. For a definition of the error see \cref{eq:error}. Note how smaller errors generally lead to higher accuracies.}
    \label{fig:equivariance}
\end{figure}

To further validate the intuition behind these models, we propose a similarity metric as an indirect measure for its equivariance, similar to \cite{sosnovik_scale-equivariant_2020,worrall_deep_2019}. In the STIR data, variance is only caused by the specific subject, the chosen position, and the chosen scale. After removing the effects of position and scale, model features and predictions should thus be invariant for each subject.

We evaluate this property within each of the three scale segments also used for the evaluation scenarios. More specifically, we compute feature map similarity between scales $64 \times 64$ and $49 \times 49$, $48 \times 48$ and $33 \times 33$ as well as $32 \times 32$ and $17 \times 17$. Feature maps are cropped to the known subject area and then interpolated to a selected target scale $s_2$ using bicubic interpolation. The relative similarity between two scales $s_1$ and $s_2$ is computed based on their feature maps $\mathbf{x}_{s_1}$ and $\mathbf{x}_{s_2}$ as
\begin{equation}\label{eq:error}
    \epsilon_{s_1 \rightarrow s_2} = \frac{\lVert \mathbf{x}_{s_1} - \mathbf{x}_{s_1} \rVert_2^2}{\lVert \mathbf{x}_{s_2} \rVert_2^2}.
\end{equation}

Equivariance errors are symmetrically computed for all scales within the testing region and averaged across all testing subjects. Specifically, for the \textit{Mid2Rest} scenario we compute $\frac{1}{4}(\epsilon_{64 \rightarrow 49} + \epsilon_{49 \rightarrow 64} + \epsilon_{32 \rightarrow 17} + \epsilon_{17 \rightarrow 32})$. \Cref{fig:equivariance} illustrates the correlation between this error and the testing accuracy on the emoji dataset for the \textit{Mid2Rest} scenario. Additional ones are provided in supplementary \cref{fig:app_equivariance}. Our evaluation is limited to the given four models as the metric requires pooled feature maps at a single scale.\footnote{Equivariance errors for the SES models can only be computed for scale changes covered by scales $s$ given in \cref{sec:reference}. These do not match those of our dataset and evaluation scenarios.}

As expected, there is a trend whereby smaller equivariance errors lead to higher accuracies when comparing different models. Notably, though, there is no such apparent correlation within each individual model. Similar results are observed on other datasets as well, although errors of $\infty$ or undefined caused by the division could not be considered.

\subsection{Scale Selection}\label{sec:indices}

\begin{figure}
    \centering
    \includegraphics[scale=0.9]{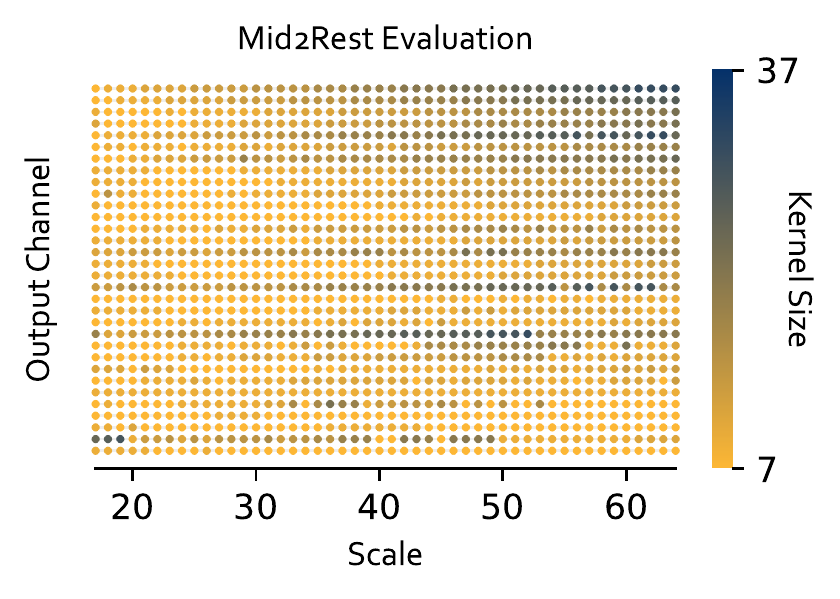}
    \caption{Scale selection by \textit{SlicePool}. Each row of points represents one of 32 output channels, with each point colored by the selected kernel size (see also \cref{fig:concept}). Correlation between input scale and kernel size is strong ($r > 0.9$) for the top ten.}
    \label{fig:indices}
\end{figure}

A key argument made when introducing our pooling methods is that properly scale-equivariant models should choose kernel sizes that are in relation to input sizes. Concretely for \cref{fig:concept}, we expect that the feature map outputs of the $3 \times 3$ kernel are selected for small input scales, and those of the $7 \times 7$ kernel are chosen for larger ones. We validate our intuition by recording the selected kernel sizes of the last layer for all input scales $s \in \{17, \ldots, 64\}$.

\Cref{fig:indices} illustrates the mentioned scale selection of the best \textit{SlicePool} model on the \textit{Mid2Rest} scenario. Each row of points represents a single output channel, with individual points being colored based on the kernel size that was selected for the smile emoji given in \cref{fig:dataset}. For each channel, we compute Pearson's correlation coefficient $r$ between input scales and selected kernel sizes. A similar representation is given in supplementary \cref{fig:app_indices} for \textit{PixelPool}.

We note that output channels with good correlations exist. In particular, ten of these have a correlation coefficient $r > 0.9$. However, the maximum kernel size used by \textit{SlicePool} is only $37 \times 37$. For \textit{PixelPool} this correlation is more difficult to measure. What we do observe is that it tends to choose larger kernels, up to $57 \times 57$, near the center, and smaller kernels towards the borders of the image.

\subsection{Computational Complexity}\label{sec:computations}

\begin{table}
    \centering
\begin{tabular}{lrrrr}
\toprule
Model & emoji & mnist & tsign & aerial \\
\midrule
Standard & 0.11s & 1.22s & 1.26s & 0.71s \\
SpatialTrans & 0.23s & 3.20s & 2.62s & 1.46s \\
Ensemble & 0.23s & 4.45s & 2.46s & 1.38s \\
Xu & 0.28s & 3.93s & 3.16s & 1.77s \\
Kanazawa & 0.27s & 3.77s & 3.34s & 1.88s \\
Hermite & 0.20s & 2.72s & 2.22s & 1.25s \\
Disco & 0.20s & 2.74s & 2.22s & 1.25s \\
PixelPool (Ours) & 6.79s & 94.61s & 80.99s & 45.57s \\
SlicePool (Ours) & 7.29s & 90.65s & 86.35s & 48.59s \\
\bottomrule
\end{tabular}
    \caption{Mean run time of the last epoch in seconds for various models and datasets on \textit{Mid2Rest} scenario. Measuring the last epoch ensures warm-up does not affect measurements.}
    \label{tab:runtime}
\end{table}

To evaluate the usability of different models and the impact on the overall training budget, both time- and resource-wise, we measure run time per epoch. The mean run time of the last epoch across all seeds is given in \cref{tab:runtime} for all models and datasets. By using the last epoch, we ensure that warm-up does not affect results disproportionally.

All models incur some additional cost compared to the \textit{Standard} CNN, in most cases by a factor of two or three. The exception to this rule are our proposed models, which are one to two orders of magnitude more expensive to compute due to the many convolutions. For example, a single convolution requires $33 \cdot 10^4$ operations, while the multiple convolutions in our layer require $38 \cdot 10^6$. This is in line with expectations, as we were interested in testing the usefulness of the method rather than prematurely optimizing it. There is potential to greatly reduce computational complexity by either merging duplicate multiplications or reducing the number of scales.

\section{Discussion}

The STIR benchmark allows us to extensively evaluate generalization to unseen scales, as it contains both trivial and realistic subsets (see \cref{fig:dataset} and \cref{sec:data}). The emojis are free of interpolation artifacts and class distribution shifts, while the traffic signs and aerial objects reflect real-world applications typically subject to changes in scale. On this benchmark, we find that no single model is superior to others (see \cref{fig:performance} and \cref{sec:performance}).

Among the models evaluated, \textit{Hermite} \cite{sosnovik_scale-equivariant_2020} and \textit{Disco} \cite{sosnovik_disco_2021} have been specifically designed to learn kernels that are scale-equivariant. In addition, we empirically show that both \textit{PixelPool} and \textit{SlicePool} improve scale equivariance over the closely related \textit{Kanazawa} \cite{kanazawa_locally_2014} and \textit{Standard} (see \cref{fig:equivariance} and \cref{sec:equivariance}). Despite that, we find that they fail to generalize well to unknown scales and are often outperformed by \textit{Ensemble} (see \cref{fig:generalization} and \cref{sec:generalization}). This indicates either that scale equivariance itself might not be a sufficient property or that these models are not as scale-equivariant as initially assumed. However, we do show that smaller equivariance errors lead to better performance (see \cref{fig:equivariance} and \cref{sec:equivariance}), although we note that this analysis is limited by the required interpolation and numerical instability for small activations. Further research is required to determine which factors contribute to good scale generalization. For example, it is known that CNNs tend to focus on textures rather than shapes \cite{geirhos_imagenet-trained_2022}, which would be influenced by scales. Also, recent findings suggest that neural networks bias toward lower frequency patterns \cite{rahaman_spectral_2019}, which may explain why generalization towards smaller scales in particular remains more difficult than towards larger scales (see \cref{fig:generalization} and \cref{sec:generalization}).

We make a few notable observations based on our experiments. The good performance (see \cref{fig:equivariance} and \cref{sec:equivariance}) and generalization (see \cref{fig:generalization} and \cref{sec:generalization}) of the \textit{Ensemble} indicates that global equivariance may be sufficient or even preferable. Our \textit{PixelPool} and \textit{SlicePool} models outperform the \textit{Kanazawa} \cite{kanazawa_locally_2014} model they are based on and the theoretically motivated SES models (see \cref{fig:generalization} and \cref{sec:generalization}), potentially due to the large number of sequential scales covered. Due to the added cost of the many associated convolutions (see \cref{sec:computations}), we had to forgo input padding, though. Our analysis of scale selection (see \cref{fig:indices} and \cref{sec:indices}) reveals that this may cause undesired boundary effects \cite{semih_kayhan_translation_2020} by the differently sized feature maps. When optimizing computations, further variations, such as those presented in supplementary \cref{sec:app_methods}, may be investigated.

\section{Conclusion}

Given the lack of an established benchmark for scale generalization, we introduce the novel Scaled and Translated Image Recognition (STIR) dataset. Based on our experiments, we find that neither the existing nor newly proposed models clearly outperform each other. This further highlights the need for a common benchmark to evaluate on. Many methods, including ours, have been shown to improve scale equivariance theoretically or empirically. Still, they fail to generalize well when faced with previously unseen scales. This suggests that either scale equivariance might not be as strong a property as originally believed or that our current ways of evaluating it are too limited.

By means of extensive evaluations we make a set of key observations that may help guide future research on the topic. First, generalization towards smaller scales is overall challenging, likely due to an inability to learn higher frequency patterns \cite{rahaman_spectral_2019}. Second, ensembles are very competitive, suggesting that scale may be more important globally rather than locally at each layer. Third, application-inspired methods such as ours are potentially subject to boundary effects \cite{semih_kayhan_translation_2020} that may break scale equivariance. We will explore these insights in future research to better understand and improve scale generalization. 


{
\small

\bibliographystyle{ieee_fullname}
\bibliography{references}

\begin{thebibliography}{10}\itemsep=-1pt

\bibitem{arcos-garcia_deep_2018}
Álvaro Arcos-García, Juan~A. Álvarez García, and Luis~M. Soria-Morillo.
\newblock Deep neural network for traffic sign recognition systems: {An}
  analysis of spatial transformers and stochastic optimisation methods.
\newblock {\em Neural Netw.}, 99:158--165, Mar. 2018.

\bibitem{boominathan_crowdnet_2016}
Lokesh Boominathan, Srinivas S.~S. Kruthiventi, and R.~Venkatesh Babu.
\newblock {CrowdNet}: {A} {Deep} {Convolutional} {Network} for {Dense} {Crowd}
  {Counting}.
\newblock In {\em Proc. 24th ACM Int. Conf. Multimedia}, pages 640--644,
  Amsterdam, Netherlands, 2016. ACM.

\bibitem{cai_unified_2016}
Zhaowei Cai, Quanfu Fan, Rogerio~S. Feris, and Nuno Vasconcelos.
\newblock A {Unified} {Multi}-scale {Deep} {Convolutional} {Neural} {Network}
  for {Fast} {Object} {Detection}.
\newblock In {\em 2016 14th Eur. Conf. Comput. Vision (ECCV)}, pages 354--370,
  Amsterdam, Netherlands, 2016. Springer.

\bibitem{chen_attention_2016}
Liang-Chieh Chen, Yi Yang, Jiang Wang, Wei Xu, and Alan~L. Yuille.
\newblock Attention to {Scale}: {Scale}-{Aware} {Semantic} {Image}
  {Segmentation}.
\newblock In {\em 2016 IEEE Conf. Comput. Vision and Pattern Recognition
  (CVPR)}, pages 3640--3649, Las Vegas, NV, USA, June 2016.

\bibitem{cohen_group_2016}
Taco Cohen and Max Welling.
\newblock Group {Equivariant} {Convolutional} {Networks}.
\newblock In {\em Proc. 33rd Int. Conf. Mach. Learn. (ICML)}, pages 2990--2999,
  New York City, NY, USA, June 2016. PMLR.

\bibitem{deng_imagenet_2009}
Jia Deng, Wei Dong, Richard Socher, Li-Jia Li, Kai Li, and Li Fei-Fei.
\newblock {ImageNet}: {A} large-scale hierarchical image database.
\newblock In {\em 2009 IEEE Conf. Comput. Vision and Pattern Recognition
  (CVPR)}, pages 248--255, Miami, FL, USA, June 2009.

\bibitem{ertler_mapillary_2020}
Christian Ertler, Jerneja Mislej, Tobias Ollmann, Lorenzo Porzi, Gerhard
  Neuhold, and Yubin Kuang.
\newblock The {Mapillary} {Traffic} {Sign} {Dataset} for {Detection} and
  {Classification} on a {Global} {Scale}.
\newblock In {\em 2020 16th Eur. Conf. Comput. Vision (ECCV)}, Glasgow, UK,
  Aug. 2020.

\bibitem{farabet_learning_2013}
Clement Farabet, Camille Couprie, Laurent Najman, and Yann LeCun.
\newblock Learning {Hierarchical} {Features} for {Scene} {Labeling}.
\newblock {\em IEEE Trans. Pattern Anal. and Mach. Intell.}, 35(8):1915--1929,
  Aug. 2013.

\bibitem{finnveden_understanding_2021}
Lukas Finnveden, Ylva Jansson, and Tony Lindeberg.
\newblock Understanding when spatial transformer networks do not support
  invariance, and what to do about it.
\newblock In {\em 2020 25th Int. Conf. Pattern Recognition (ICPR)}, pages
  3427--3434, Milan, Italy, Jan. 2021.

\bibitem{fontawesome_download_2022}
Dave Gandy, Jason Otero, Edward Emanuel, Frances Botsford, Jason Lundien,
  Kelsey Jackson, Mike Wilkerson, Rob Madole, Jory Raphael, Travis Chase,
  Geremia Taglialatela, Brian Talbot, and Trevor Chase.
\newblock {Font} {Awesome}.
\newblock \url{https://fontawesome.com/v5/download}, Nov. 2022.

\bibitem{geirhos_imagenet-trained_2022}
Robert Geirhos, Patricia Rubisch, Claudio Michaelis, Matthias Bethge, Felix~A.
  Wichmann, and Wieland Brendel.
\newblock {ImageNet}-trained {CNNs} are biased towards texture; increasing
  shape bias improves accuracy and robustness.
\newblock In {\em 2019 Int. Conf. on Learn. Representations (ICLR)}, New
  Orleans, LA, USA, May 2019.

\bibitem{ghosh_scale_2019}
Rohan Ghosh and Anupam~K. Gupta.
\newblock Scale {Steerable} {Filters} for {Locally} {Scale}-{Invariant}
  {Convolutional} {Neural} {Networks}.
\newblock {\em arXiv:1906.03861 [cs]}, June 2019.

\bibitem{hinton_matrix_2018}
Geoffrey~E. Hinton, Sara Sabour, and Nicholas Frosst.
\newblock Matrix capsules with {EM} routing.
\newblock In {\em 2018 Int. Conf. Learn. Representations (ICLR)}, Vancouver,
  BC, Canada, Feb. 2018.

\bibitem{jaderberg_spatial_2015}
Max Jaderberg, Karen Simonyan, Andrew Zisserman, and koray kavukcuoglu.
\newblock Spatial {Transformer} {Networks}.
\newblock In {\em Adv. Neural Inf. Process. Syst. (NIPS) 28}, pages 2017--2025.
  Curran Associates, Inc., Montreal, QC, Canada, 2015.

\bibitem{jansson_exploring_2021}
Ylva Jansson and Tony Lindeberg.
\newblock Exploring the ability of {CNN}s to generalise to previously unseen
  scales over wide scale ranges.
\newblock In {\em 2020 25th Int. Conf. Pattern Recognition (ICPR)}, pages
  1181--1188, Milan, Italy, Jan. 2021.

\bibitem{jansson_scale-invariant_2022}
Ylva Jansson and Tony Lindeberg.
\newblock Scale-{Invariant} {Scale}-{Channel} {Networks}: {Deep} {Networks}
  {That} {Generalise} to {Previously} {Unseen} {Scales}.
\newblock {\em J. Math. Imag. and Vision}, 64:506--536, Apr. 2022.

\bibitem{kanazawa_locally_2014}
Angjoo Kanazawa, Abhishek Sharma, and David Jacobs.
\newblock Locally {Scale}-{Invariant} {Convolutional} {Neural} {Networks}.
\newblock {\em arXiv:1412.5104 [cs]}, Dec. 2014.

\bibitem{kingma_adam_2017}
Diederik~P. Kingma and Jimmy Ba.
\newblock Adam: {A} {Method} for {Stochastic} {Optimization}.
\newblock In {\em 2015 Int. Conf. on Learn. Representations (ICLR)}, San Diego,
  CA, USA, Jan. 2017.

\bibitem{lecun_gradient-based_1998}
Yann Lecun, Léon Bottou, Yoshua Bengio, and Patrick Haffner.
\newblock Gradient-based learning applied to document recognition.
\newblock {\em Proc. IEEE}, 86(11):2278--2324, Nov. 1998.

\bibitem{liu_efficient_2018}
Yan Liu, Qirui Ren, Jiahui Geng, Meng Ding, and Jiangyun Li.
\newblock Efficient {Patch}-{Wise} {Semantic} {Segmentation} for
  {Large}-{Scale} {Remote} {Sensing} {Images}.
\newblock {\em Sensors}, 18(10):3232, Oct. 2018.

\bibitem{marcos_scale_2018}
Diego Marcos, Benjamin Kellenberger, Sylvain Lobry, and Devis Tuia.
\newblock Scale equivariance in {CNNs} with vector fields.
\newblock {\em arXiv:1807.11783 [cs]}, July 2018.

\bibitem{mouton_stride_2020}
Coenraad Mouton, Johannes~C. Myburgh, and Marelie~H. Davel.
\newblock Stride and {Translation} {Invariance} in {CNNs}.
\newblock In {\em 2021 South Afr. Conf. Artif. Intell. Res. (SACAIR)}, pages
  267--281, Muldersdrift, South Africa, Feb. 2020. Springer.

\bibitem{nam_psi-cnn_2018}
Gi~Pyo Nam, Heeseung Choi, Junghyun Cho, and Ig-Jae Kim.
\newblock {PSI}-{CNN}: {A} {Pyramid}-{Based} {Scale}-{Invariant} {CNN}
  {Architecture} for {Face} {Recognition} {Robust} to {Various} {Image}
  {Resolutions}.
\newblock {\em Appl. Sci.}, 8(9):1561, Sept. 2018.

\bibitem{rahaman_spectral_2019}
Nasim Rahaman, Aristide Baratin, Devansh Arpit, Felix Draxler, Min Lin, Fred
  Hamprecht, Yoshua Bengio, and Aaron Courville.
\newblock On the {Spectral} {Bias} of {Neural} {Networks}.
\newblock In {\em Proc. 36th Int. Conf. Mach. Learn. (ICML)}, pages 5301--5310,
  Long Beach, CA, USA, May 2019. PMLR.

\bibitem{russakovsky_imagenet_2015}
Olga Russakovsky, Jia Deng, Hao Su, Jonathan Krause, Sanjeev Satheesh, Sean Ma,
  Zhiheng Huang, Andrej Karpathy, Aditya Khosla, Michael Bernstein,
  Alexander~C. Berg, and Li Fei-Fei.
\newblock {ImageNet} {Large} {Scale} {Visual} {Recognition} {Challenge}.
\newblock {\em Int. J. Comput. Vision}, 115(3):211--252, Dec. 2015.

\bibitem{semih_kayhan_translation_2020}
Osman Semih~Kayhan and Jan~C. van Gemert.
\newblock On {Translation} {Invariance} in {CNNs}: {Convolutional} {Layers}
  {Can} {Exploit} {Absolute} {Spatial} {Location}.
\newblock In {\em 2020 IEEE/CVF Conf. Comput. Vision and Pattern Recognition
  (CVPR)}, pages 14262--14273, Seattle, WA, USA, June 2020.

\bibitem{sermanet_traffic_2011}
Pierre Sermanet and Yann LeCun.
\newblock Traffic sign recognition with multi-scale {Convolutional} {Networks}.
\newblock In {\em 2011 Int. Joint Conf. Neural Netw. (IJCNN)}, pages
  2809--2813, San Jose, CA, USA, July 2011.

\bibitem{shen_multi-scale_2015}
Wei Shen, Mu Zhou, Feng Yang, Caiyun Yang, and Jie Tian.
\newblock Multi-scale {Convolutional} {Neural} {Networks} for {Lung} {Nodule}
  {Classification}.
\newblock In {\em Inf. Process. Med. Imag.}, pages 588--599, Sabhal Mor Ostaig,
  Isle of Skye, UK, 2015. Springer.

\bibitem{sosnovik_disco_2021}
Ivan Sosnovik, Artem Moskalev, and Arnold Smeulders.
\newblock {DISCO}: accurate {Discrete} {Scale} {Convolutions}.
\newblock In {\em 2021 32nd Brit. Mach. Vision Conf. (BMVC)}, Virtual Event,
  UK, Oct. 2021.

\bibitem{sosnovik_scale-equivariant_2020}
Ivan Sosnovik, Michał Szmaja, and Arnold Smeulders.
\newblock Scale-{Equivariant} {Steerable} {Networks}.
\newblock In {\em 2020 Int. Conf. Learn. Representations (ICLR)}, Virtual
  Event, USA, Apr. 2020.

\bibitem{van_noord_learning_2017}
Nanne van Noord and Eric Postma.
\newblock Learning scale-variant and scale-invariant features for deep image
  classification.
\newblock {\em Pattern Recognition}, 61:583--592, Jan. 2017.

\bibitem{worrall_deep_2019}
Daniel Worrall and Max Welling.
\newblock Deep {Scale}-spaces: {Equivariance} {Over} {Scale}.
\newblock In {\em Adv. Neural Inf. Process. Syst. (NeurIPS) 32}. Curran
  Associates, Inc., Vancouver, BC, Canada, 2019.

\bibitem{xia_dota_2018}
Gui-Song Xia, Xiang Bai, Jian Ding, Zhen Zhu, Serge Belongie, Jiebo Luo, Mihai
  Datcu, Marcello Pelillo, and Liangpei Zhang.
\newblock {DOTA}: {A} {Large}-{Scale} {Dataset} for {Object} {Detection} in
  {Aerial} {Images}.
\newblock In {\em 2018 IEEE/CVF Conf. Comput. Vision and Pattern Recognition
  (CVPR)}, pages 3974--3983, Salt Lake City, UT, USA, June 2018.

\bibitem{xu_scale-invariant_2014}
Yichong Xu, Tianjun Xiao, Jiaxing Zhang, Kuiyuan Yang, and Zheng Zhang.
\newblock Scale-{Invariant} {Convolutional} {Neural} {Networks}.
\newblock {\em arXiv:1411.6369 [cs]}, Nov. 2014.

\bibitem{zhou_scale-transferrable_2018}
Peng Zhou, Bingbing Ni, Cong Geng, Jianguo Hu, and Yi Xu.
\newblock Scale-{Transferrable} {Object} {Detection}.
\newblock In {\em 2018 IEEE/CVF Conf. Comput. Vision and Pattern Recognition
  (CVPR)}, pages 528--537, Salt Lake City, UT, USA, June 2018.

\bibitem{zhu_scaling-translation-equivariant_2021}
Wei Zhu, Qiang Qiu, Robert Calderbank, Guillermo Sapiro, and Xiuyuan Cheng.
\newblock Scaling-{Translation}-{Equivariant} {Networks} with {Decomposed}
  {Convolutional} {Filters}.
\newblock {\em arXiv:1909.11193 [cs, stat]}, May 2021.

\end{thebibliography}
}

\newpage
\onecolumn

\appendix
\section{Model Performance}\label{sec:app_performance}

\begin{figure*}[h!]
    \centering
    \includegraphics[scale=0.9]{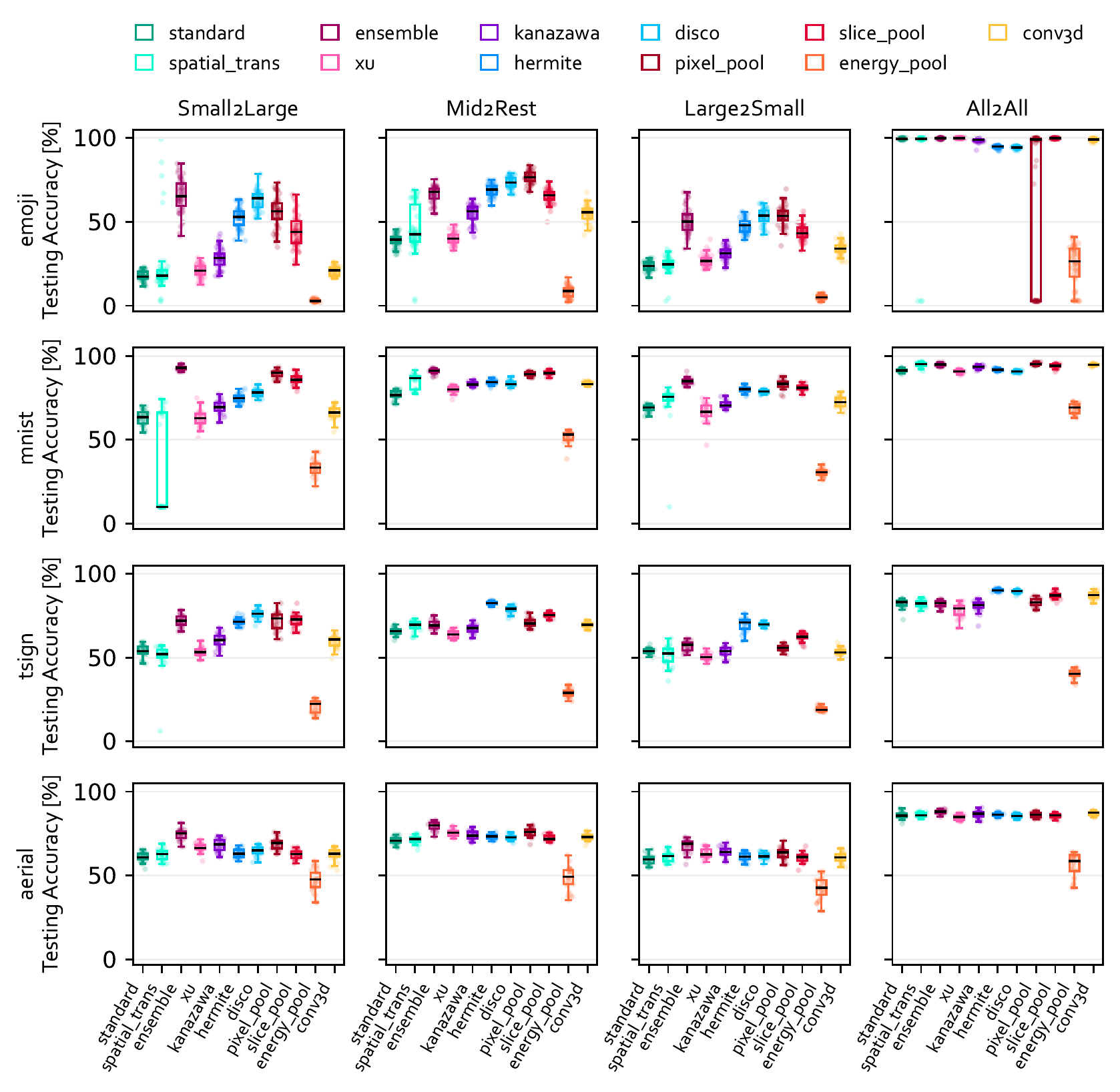}
    \caption{Detailed plots of model performances across different datasets and evaluation scenarios. Rows correspond to emoji, mnist, tsign and aerial datasets (from top to bottom). Columns correspond to \textit{Small2Large}, \textit{Mid2Rest}, \textit{Large2Small} and \textit{All2All} evaluation scenarios (from left to right). Box plots summarize the median and quartiles of individual model performances, disregarding outliers. Scatter plots indicate individual model performances, whereby 50 trials were performed for the emoji dataset and 25 for others. Note how in some cases, performances are highly dependent on the seed used for initialization. Descriptions of energy\_pool and conv3d are given in Sec.~\ref{sec:app_methods}.}
    \label{fig:app_performance}
\end{figure*}

\newpage

\section{Scale Generalization}\label{sec:app_generalization}

\begin{figure*}[h!]
    \centering
    \includegraphics[scale=0.9]{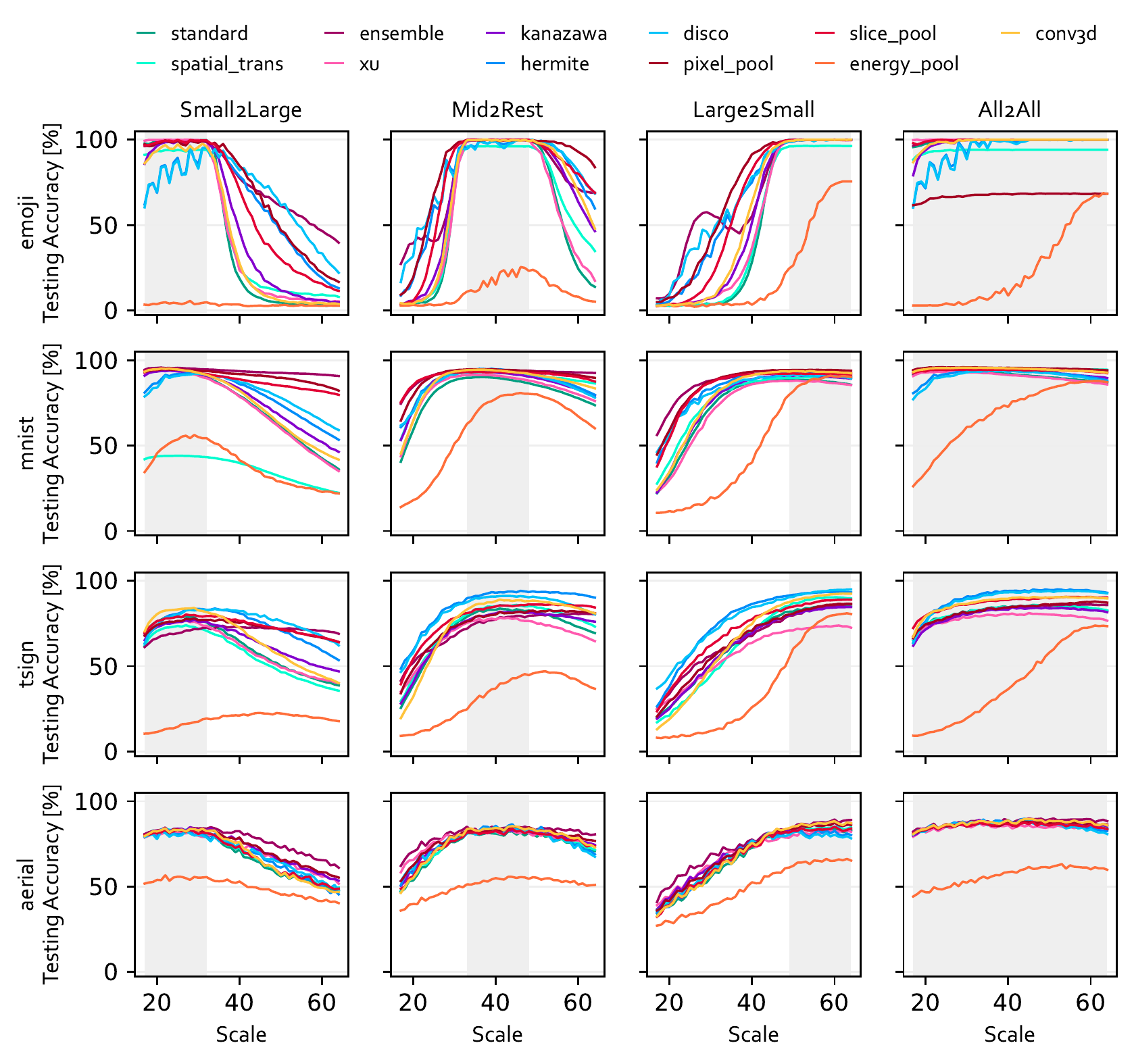}
    \caption{Detailed plots of scale generalization across different datasets and evaluation scenarios. Rows correspond to emoji, mnist, tsign and aerial datasets (from top to bottom). Columns correspond to \textit{Small2Large}, \textit{Mid2Rest}, \textit{Large2Small} and \textit{All2All} evaluation scenarios (from left to right). Lines illustrate scale-dependent performance and follow the mean of 50 repetitions on the emoji dataset and 25 for others. Shaded areas represent scales used during training and thus regions without a data distribution shift. Descriptions of energy\_pool and conv3d are given in Sec.~\ref{sec:app_methods}.}
    \label{fig:app_generalization}
\end{figure*}

\newpage

\section{Feature Map Similarity}\label{sec:app_equivariance}

\begin{figure*}[h!]
    \centering
    \includegraphics[scale=0.9]{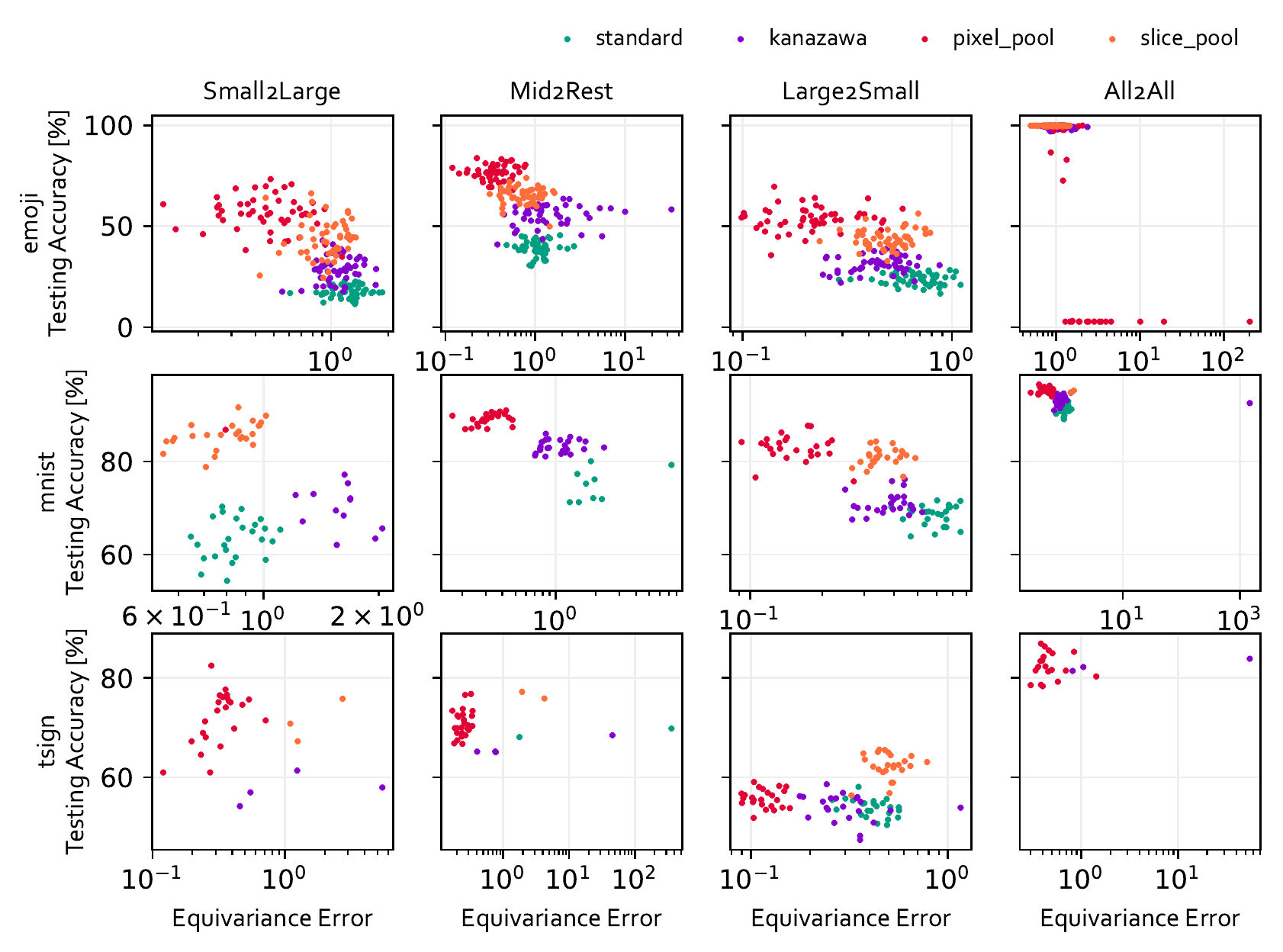}
    \caption{Measurement of equivariance errors versus testing accuracy across different datasets and evaluation scenarios. Rows correspond to emoji, mnist and tsign datasets (from top to bottom). Columns correspond to \textit{Small2Large}, \textit{Mid2Rest}, \textit{Large2Small} and \textit{All2All} evaluation scenarios (from left to right). An evaluation for aerial is missing as it incorporates non-square objects that complicate its assessment. Missing points are caused by division by zero, which leads to errors of $\infty$ or undefined that cannot be plotted.}
    \label{fig:app_equivariance}
\end{figure*}

\newpage

\section{Scale Selection}\label{sec:app_indices}

\begin{figure*}[h!]
    \centering
    \includegraphics[scale=0.9]{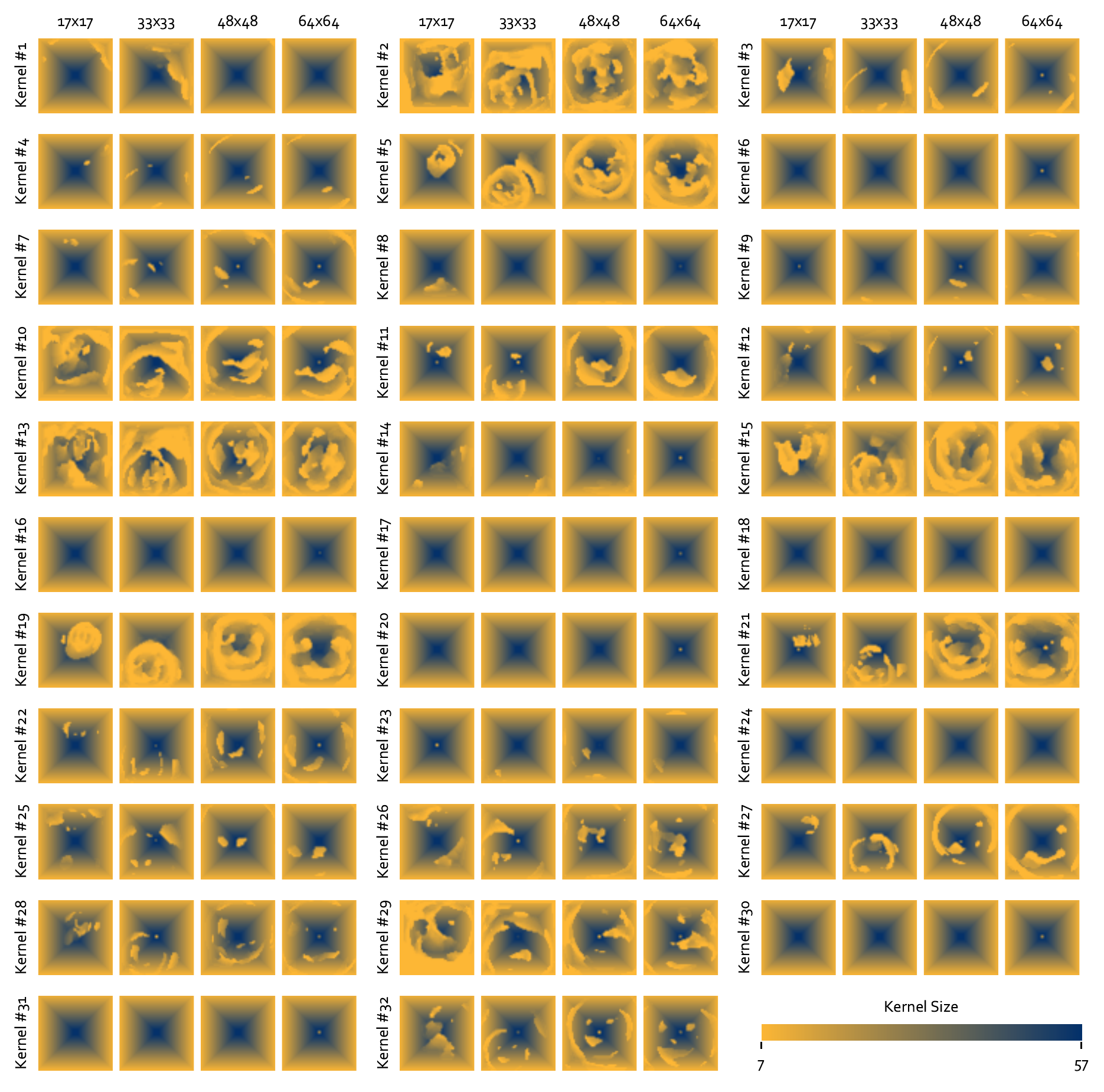}
    \caption{Analysis of scale selection for \textit{PixelPool}. Kernel numbers refer to the $n=32$ kernels or output channels of the last \textit{PixelPool} convolutional layer. Colors indicate the kernel size chosen by the maximum pooling for each pixel. We only illustrate the response to the smile emoji for scales $s \in \{17, 33, 48, 64\}$ for each kernel. Note how the default selection uses larger (blue) kernel sizes at the center and smaller (yellow) ones at the borders.}
    \label{fig:app_indices}
\end{figure*}

\newpage

\section{Energy Pooling and Three-Dimensional Convolutions}\label{sec:app_methods}

In addition to the methods discussed in the paper, the following methods were investigated to further process the scale domain. However, due to inherent limitations they are not included in the final results.

\textbf{EnergyPool} first computes the sum of all activations across spatial pixels $(i_x, i_y)$ at each different scale. Afterwards, the scale $i_s$ that corresponds to the maximum sum is chosen. The output then again consists of all values at positions $(j_s, j_x, j_y)$ where $i_s = j_s$. One can think of this as the average pooling variant of the \textit{SlicePool} presented in the paper. However, it is inherently biased towards small kernels without additional adaptions to handle zero-padding at different scales. 

\textbf{Conv3d} uses a simple three-dimensional convolution instead of pooling. This allows the model to learn correlations among all three dimensions, including the scale. There are, however, two major limitations to this approach. First, the operation itself does not attempt to be scale-equivariant. Secondly, it maintains the scale dimension itself. We thus only consider a model that directly applies global pooling.

\section{Computation of Operations}\label{sec:app_computations}

Operations are either additions, subtractions, multiplications, divisions, or maximums. Let $n$ be the height and width of the input image or feature map and $k$ the kernel size. Then the number of operations of a standard convolutional layer is given by
\begin{equation}
    o_c(k) = 2 \cdot k^2 \cdot (n - k + 1)^2.
\end{equation}

For the scaled convolutional layer, $s$ convolutions are performed, where $s$ is the number of scales given in Eq.~(3) of the paper. Each scale uses a different kernel size, such that the number of operations is, without further optimizations, given by
\begin{equation}
    o_s(k) = \sum_{i=0}^{s-1} o_c(k + 2 \cdot i).
\end{equation}

\begin{figure}[h!]
    \centering
    \includegraphics[scale=0.9]{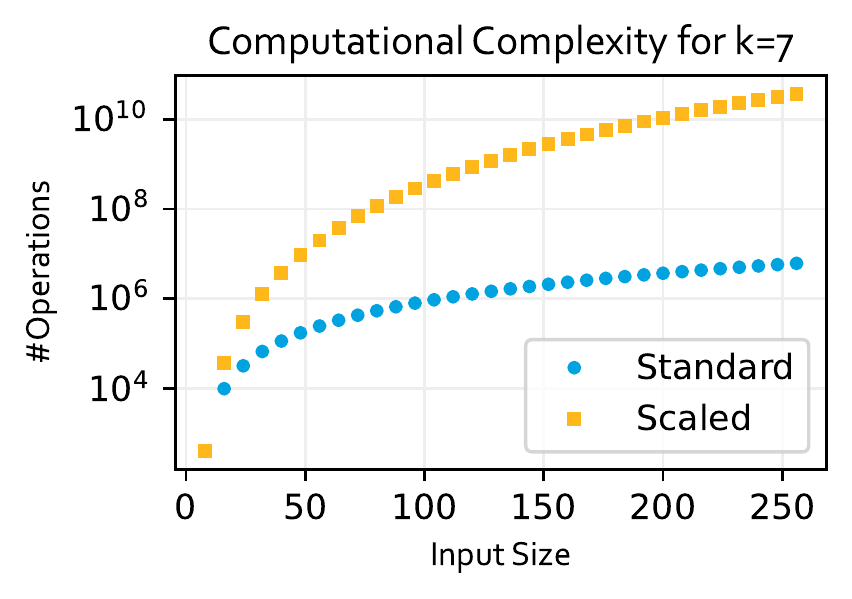}
    \caption{Number of operations for kernel size $k=7$. For typical image sizes, the scaled convolutional layer requires several orders of magnitudes more operations than the standard.}
    \label{fig:operations}
\end{figure}

Other operations caused by the interpolation and pooling would also need to be factored in, but have comparatively little impact. Fig.~\ref{fig:operations} exemplifies the complexity for different input sizes with a fixed kernel size of $k=7$. It is evident that for typical image sizes, the computational load is several orders of magnitude higher when comparing scaled and standard convolutional layers.

\end{document}